# An effective coaxiality measurement for twist drill based on line structured light sensor

Ailing Cheng, Jiaojiao Ye, Fei Yang, Shufang Lu, Fei Gao

*Abstract*—Aiming at the accurate and effective coaxiality measurement for twist drill with irregular surface, an optical measurement mechanism is proposed in this paper. First, A high-precision rotation instrument based on four core units is designed, which can obtain the 3-D point cloud data of full angle for the twist drill. Second, in the data processing stage, an improved robust Gaussian mixture model is established for accurate and rapid blade back segmentation. To improve measurement efficiency, a rapid reconstruction method of the twist drill axis based on orthogonal synthesis is provided to locate the axial position of the maximum deviation from the benchmark by utilizing the extracted blade back data. Finally, by calculating the maximum radial Euclidean distance from the benchmark, the coaxiality error of the twist drill is obtained. Comparing with other measurement methods, experimental results show that our proposed method is effective with high precision of 3 um and high efficiency of less than 3 s/pc. The result demonstrate that the proposed method is effective, robust and automatic, it can be applied in many actual industrial scene.

## I. Introduction

Twist drill is an important mechanical tool for drilling round holes in workpiece. It consists of a handle part and a working part, as shown in Fig. 1. As an important tolerance of the rotating body, the coaxiality directly affects the processing quality [1]. Twist drills with large coaxiality errors will cause hole size deviation and damage the machined workpiece. Therefore, in the intelligent manufacturing environment, it is of great significance to quickly and accurately measure the coaxiality error of twist drills.

Despite the rapid development of different measurement techniques, online precision measurement of complex surfaces are still facing challenging problems [2]. At present, common contact coaxiality measurement technologies include bearing gauge, coaxiality measurement instrument, and CMM (coordinate measuring machine) [3] [4] [5] Among them, the CMM [6] equipped with a certain precision scanning probes can perform high-precision and robust measurements [2]. However, like other contact methods, due to the complicated operation and slow measurement process, it is difficult to be applied in automatic manufacturing scenarioes.

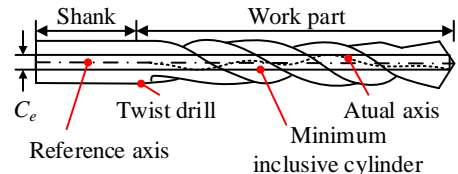

**Fig. 1.** Illustration of coaxiality for twist drill. $C_e$ is the coaxiality of the twist drill

With the rapid development of machine vision and optics [7] [8] [9] [10] [11], photoelectric has become an attractive technology in non-contact measurement methods. They can have wide measurement range and dense sampling rate [2]. At present, non-contact optical measurement technologies can be divided into two categories: the passive and the active methods. Passive methods use stereo-vision technology [12] [13] to reconstruct 3D topography of object surfaces without active illumination. However, due to the need to detect corresponding different images pairs, the measurement accuracy varies with the surface texture of the object. The time-of-flight (TOF) technique [14] uses the time-of-flight of a signal to measure the distance between asynchronous transceivers (or reflected surfaces). The entire system can be very compact and they are suitable for mobile applications. However, this technology still suffers from low image resolution, high power consumption and so on. The structured light technology [15] actively projects structured light with encoded information onto the surface of an object, and reconstructs the object by decoding the information. Among them, image registration is a key step and will affect the measurement accuracy. Line structured light vision technology is an effective approach, and can monitor the shape of a 3-D revolving-symmetry by reconstructing normal section profiles [16]. The technology enables high-precision measurements by employing high-power lasers and high-resolution cameras. Therefore, combined with a certain rotating structure, the measuring mechanism can be made very flexible and precise. Inspired by this, a coaxiality measurement mechanism based on line structured light sensor is proposed to measure twist drills.

At present, non-contact measurement technology has been studied and applied in several industrial manufacturing scenarios [3] [7] [8] [17], such as object inspection, surface measurement of complex parts, dimensional tolerance and shape tolerance measurement of parts. Sun *et al.* [16] proposed

This work is being supported by the National Key Research and Development Project of China under Grant No. 2020AAA0104001, the Zhejiang Lab. under Grant No. 2019KD0AD011005 and the Zhejiang Provincial Science and Technology Planning Key Project of China under Grant No. 2021C03129.

Ailing Cheng, Shufang Lu, and Fei Gao are with the Laboratory of Graphics & Image Processing, College of Computer Science and Technology, Zhejiang University of Technology, Hangzhou, Zhejiang Province 310023, China (e-mail: vinvia2008@163.com; sflu@zjut.edu.cn; feig@zjut.edu.cn). Jiaojiao Ye and Fei Yang are with the Research Center of Intelligent Computing Software, Zhejiang Lab, Hangzhou, Zhejiang Province 311121, China (e-mail: jiaojiaoye2016@gmail.com; yangf@zhejianglab.com).

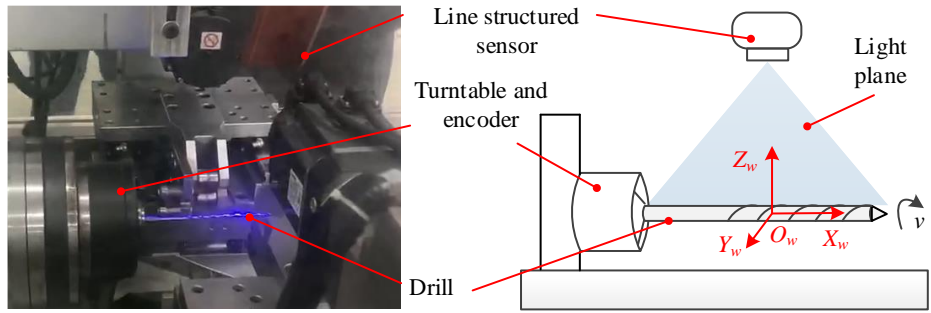

**Fig. 2.** Coaxiality Measurement Equipment for Twist Drill

a pose-unconstrained normal section profile reconstruction framework for 3-D revolving-symmetry structures. Wang *et al.* [18] adopted structured-light vision to dynamically measure rail profile, and propose a simple and effective distortion rectifying method to avoid distorted rail profiles. Li *et al.* [19] developed a novel system based on structured light binocular vision to obtain the full profile of rail and achieve more accurate and efficient rail wear measurement on site. Chai *et al.* [3] used a laser displacement sensor to measure the coaxiality of composite gears by acquiring cross-sectional contours, separating the apex data of gears and fitting the center of the cross section. Pei *et al.* [20] applied a single laser displacement sensor to realize radial jump variable measurement of gears, and improved the measurement accuracy by optimizing the laser angle and installation position. Zhang *et al.* [21] used a laser displacement sensor to obtain the surface profile of the part and fit the center of the circle to calculate the concentricity of large forging, thereby improving the operation accuracy of disassembly and assembly. Guo *et al.* [22] put forward a 3D (three-dimensional) point cloud measurement system based on a line structure light sensor to obtain information on the 3D shape of the gear tooth flank.

Even more, in addition to high-resolution data acquisition, the precise measurement of dimensional micro-features also needs tailor-made evaluation methods [23]. In the field of instrumentation and measurement, point cloud is one of the most primitive 3-D data representations and can accurately reflect the real size and shape structure of the object [24]. To obtain the data of target area, partitioning or reconstructing of dimensional measurement data is a crucial step. The methods can be divided into several categories, such as edge-based, region-growing, model-fitting, attribute clustering, and hybrid approaches [23][25] [26] [27].

Edge-based methods perform segmentation by detecting different metrics on neighboring pixels and use them as transition zones between segmented elements. This method is fast, but also very sensitive to noisy data [23] [27]. Region-growing methods combine adjacent pixel points with seed points according to certain criterion [24], but these methods tent to under or over segmentation [23] [24] [27]. Zhang *et al.* [24] proposed a region growing method based on 2-D-3-D mutual projections, which divides the visible points and the occlusion points by selecting initial seed points with the geometric information. Ma *et al.* [28] proposed a segmentation framework based on region-growing method including neighborhood search, filter sampling, Euclidean clustering, which can improve the speed and accuracy of the algorithm. Model-based methods utilize geometric primitives (such as planes, spheres, or cylinders) to group the measured points [23] [25]. This method is widely used for partitioning tasks in reverse engineering. Inspired by the model-based idea, Erdenebayar *et al.* [25] put forward flake surface recognition model that can remove noisy point clouds by resampling. In [29], Zhang *et al.* presented to integrate normal-angle cues into discriminative feature learning to enhance local structure representation for small objects. Attribute clustering method consists of attribute computation and attribute-based clustering of point clouds. In either way, for complex surface measurement, any method needs the ability to automatically and optimally partition micro-features data with low uncertainty, so as to obtain maximum number of acquired points to associate with the corresponding geometric element [30]. It is proved that hybrid approaches demonstrably have the potential for an automated optimal partitioning [23]. Lübke *et al.* [30] demenstrated that the automated partitioning was little sensitive regarding the initial values of the approximation and can converge reliably. Freyberg *et al*. [31] successfully applied the method with automated partitioning in micro-measurements evaluation. In addition, in [32] a statistical method is combined for automatic detection of outliers. In the coaxiality measurement of the twist drill, the segmentation algorithm needs the ability to automatically partitioning aquired data with micro-change of surface normal-angle cues of different geometric elements. Inspired by the above algorithms, an automated partitioning idea is integrated into the attribute clustering method. In this idea, a statistical method based on our improved GMM (Gaussian Mixture Modeling) is designed to learn the discriminative features of geometric elements on the twist drill. In the our improved GMM model, local spatial neighboring information is introduced to enhance the micro-features of each element such as blade back and blade groove, and overcome the sensitivity of the classical GMM to noise.

In all, to improve the efficiency and accuracy of coaxiality measurement for twist drill, first, an effective measurement mechanism based on line structured light sensor is proposed, which can obtain the full angle data of the twist drill. Second, thanks to the worthy development of unsupervised machine learning, GMM can be used to fit arbitrary probability density function with strong approximation ability and high robustness [33]. It is widely used in object inspecting [33], background modeling [34], modeling segmentation model [35] and other fields. Aiming to partition geometric elements of the twist drill and obtain measurement data on circular cross section for coaxiality calculation, an improved GMM-based segmentation method is proposed to learn the discriminative features distribution of geometric elements. A local (neighborhood) spatial information is introduced to overcome the sensitivity of the classical GMM to noise. Third, to improve efficiency, a rapid axis reconstruction method based on orthogonal synthesis is presented to locate the maximum deviation of the actual axis from the benchmark. It takes less than 3 s to obtain the whole profile data of a twist drill and the measurement standard deviation is within 0.007 mm. The coaxiality error of a twist drill can be measured in 3 s using the measurement method proposed in this paper. It can realize the flexible measurement of different specifications (within a certain range) of twist drills. The main contributions are as following:

1) A 3D coaxiality measurement mechanism for twist drills is proposed, which contains three core modules such as: three-rule system calibration, blade back extraction, and maximum deviation location and calculation of the axis. The proposed mechanism can acquire full angle 3D data of the twist drill by rotation for precision measurement.
2) An improved GMM-based segmentation method is investigated to learn discriminative features of geometric elements and segment the point cloud data. In which, a local spatial information is introduced to overcome the sensitivity of the classical GMM to noise, and a two-level division strategy is designed to construct local neighborhood. Experiment results demonstrate that the investigated method is robust and accurate.
3) A rapid reconstruction method of axis based on orthogonal synthesis is provided to locate the axial position of the cross section with the maximum radial deviation. In this method, the depth differences of arbitrary two groups axisymmetric outlines are calculated, and then the actual axis can be approximated by orthogonal synthesis of these two differences. The method is accurate and rapid, and can easily handle the problem of the data deficiency caused by the excessive bending of the twist drill.

The remainder of this article is organized as follows. Section II describes the designed structures of our measurement system. Section III illustrates the proposed segmentation method of the blade back and the reconstruction method of the axis for the twist drill. We detail our experimental results and analysis in Section IV. Finally, Section V concludes this article.

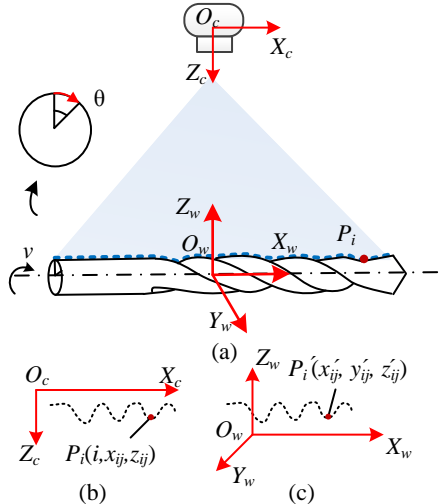

**Fig. 3.** (a) Illustration of point cloud collection. (b) Point cloud in sensor coordinate system. (c) Point cloud in measurement coordinate system. $P_i'$ corresponds to $P_i$.

## II. MECHANISM

### A. Components and Principle

To improve the efficiency and accuracy of coaxiality measurement for twist drill, a novel mechanism based on 3D measurement is designed, as shown in Fig. 2. The mechanism comprises several main components such as PLC, differential encoder, line structured light sensor and high-precision turntable. Among them, the turntable drives drills to run around turntable axis in command of PLC, meanwhile differential encoder acquires sequential angle signals and triggers line structured light sensor to catch point cloud data of drill surface, which are the core principle of the mechanism. On this basis, raw data is collected for blade back extraction and other following algorithms.

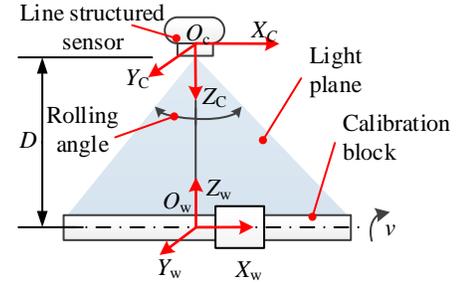

**Fig. 4.** The relationship between sensor coordinate and measurement coordinate. $O_CX_CY_CZ_C$ is the sensor coordinate system and $O_WX_WY_WZ_W$ is the measurement coordinate system.

### B. Data Acquisition

In the data acquisition phase, the drill is grasped and rotates around turntable axis in command of PLC, differential encoder acquires sequential rotation signals from the turntable, and triggers the sensor to catch point cloud data of continuous contours on the drill surface. During the process, any $i$-th sampled point data of rotating $\theta$ degree, $P_i=\{i, x_{ij}, z_{ij}\}$, can be transformed to a 3-D point data $P_i'=\{x_{ij}', y_{ij}', z_{ij}'\}$ from the sensor coordinate system to the measurement coordinate system. Where, $i=1, 2, ..., I, j=1, 2, ..., J$, $I$ is the sensor triggered times, $J$ is the point cloud number of a single sampling, as shown in Fig. 3. When the rotation reaches a cycle, the turntable is commanded to stop by PLC and the coaxiality calculation officially begins.

### C. Coordinate Transformation

As shown in Fig. 4, in the measurement system, $O_CX_CY_CZ_C$ is the sensor coordinate system, $O_WX_WY_WZ_W$ is the measurement coordinate system. The conversion from the sensor coordinate system to the measurement coordinate system is:

$$\begin{cases} \theta_i = \dfrac{360°}{I} * i \\ x_{ij}' = x_{ij} \\ y_{ij}' = (D - z_{ij}) * \sin\theta_i \\ z_{ij}' = (D - z_{ij}) * \cos\theta_i \end{cases} \quad (1)$$

Where, $i=1, 2, ..., I$, $I$ indicates the times that the sensor is triggered, $j=1, 2, ..., J$, $J$ is the number of point clouds for a single frame contour. $x_{ij}$ and $z_{ij}$ are point cloud data of the twist drill after rotating $\theta_i$ degree acquired by line structured light sensor. $x'_{ij}$, $y'_{ij}$, and $z'_{ij}$ are the transformed data corresponding $x_{ij}$, $z_{ij}$ and $\theta_i$. $D$ is the distance from the sensor to the axis of the turntable, which is the system parameter and needed to be calibrated.

*D. Calibration and Adjustment*

To determine the system parameter $D$, a calibration block with a stepped shaft is designed to calibrate the system as shown in Fig. 5. Because the surface of the twist drill is twined by spiral grooves, it may not be completely scanned according to the optical measurement constraints of line structured light sensor based on triangulation. Thus, the ladder on the calibration block can help with the sensor installation, and make the sensor achieve a relatively perfect posture and position. The pose of the sensor can be described as Euler angles including pitching angle, rolling angle and yaw angle. In the measurement system, as shown in Fig. 4, they are respectively corresponding to the rotation around $O_W X_W$, $O_W Y_W$ and $O_W Z_W$ axes.

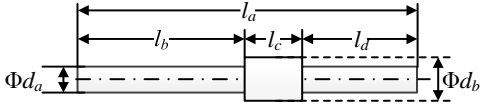

**Fig. 5.** Calibration block diagram.

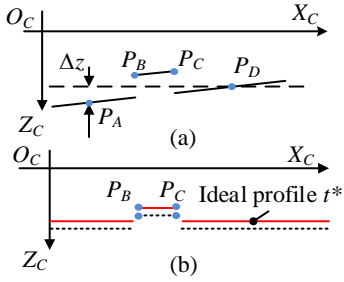

**Fig. 6.** Schematic diagram of calibration block contour. (a) is the situation in criterion I, but does not reach to criterion II and III. The red line in (b) is the situation satisfied with the criterion I, II and III. $P_A$, $P_B$, $P_C$ and $P_D$ are evenly axial spaced points, $P_B$ and $P_C$ are the two endpoints of the ladder.

According to the adjustment and calibration rules from coarse to fine, a cyclic adjustment method based on three degrees of freedom is proposed. In detail, in the mechanism installation, the installation pose and position of the sensor are continuously being fine-tuned according to adjusting installation height, rolling angle and the position on $O_W Y_W$ axis until it meets RULE I, II and III.

RULE I. The calibration block appears completely in the field of the sensor view.

RULE II. The outline of the calibration block is close to a straight line and satisfies (2).

$$\begin{cases} x_b - x_a = x_c - x_b = x_d - x_c \\ \Delta z = |z_d - z_a| \\ \Delta z \leq \Delta z_{th} \end{cases} \quad (2)$$

Where, as shown in Fig. 6(a), $(x_a, z_a)$, $(x_b, z_b)$, $(x_c, z_c)$ and $(x_d, z_d)$ are the coordinates of $P_A$, $P_B$, $P_C$ and $P_D$ collected by the sensor, respectively. $P_B$ and $P_C$ are the two endpoints of the ladder. $\Delta z_{th}$ is the line threshold given by experience and is usually set to be less than 3 times $z$ resolution.

RULE III. Fine-tune the position of the sensor on the $O_W Y_W$ axis. As shown in Fig. 6(b), for the points $P_B$ and $P_C$ on the calibration block, a series points data can be acquired and be recorded as two sets, i.e. $\{i, x_{Bi}, z_{Bi}\}$ and $\{i, x_{Ci}, z_{Ci}\}$ for $P_B$ and $P_C$, respectively. When the data satisfying (3):

$$\begin{cases} z_{Bi} + z_{Ci} < z_{B(i-1)} + z_{C(i-1)} \\ z_{Bi} + z_{Ci} < z_{B(i+1)} + z_{C(i+1)} \end{cases} \quad (3)$$

At this time, the position of the sensor on the $O_W Y_W$ axis is the best, that is, the sensor is closest to the calibration block, and $i$ is denoted as $i^*$. Then system parameter $D$ can be obtained from (4):

$$D = (z_{Bi^*} + z_{Ci^*})/2 \quad (4)$$

## III. FRAMEWORK AND METHOD

The flow chart of our measurement framework for twist drill is shown in Fig. 7.

*A. Data Preprocessing*

The data preprocessing provides a convenient data model for the blade back extraction. As shown in Fig. 8, $S$ and $P$ are the points respectively collected at the $a$-th and $b$-th time, $S''$ and $P''$ are the transformed points calculated according to (5).

$$\begin{cases} x''_{ij} = x_{ij} \\ y''_{ij} = 2\pi\gamma * \dfrac{i}{I} \\ z''_{ij} = z_{ij} \end{cases} \quad (5)$$

Where, $\gamma$ is the transformation coefficient and is given manually. $i=1, 2, ..., I, j=1, 2, ..., J$. $i$ is the number of times that the sensor was triggered. $J$ is the point clouds number of a single sampling. $\{i, x_{ij}, z_{ij}\}$ is the $i$-th sampled point data set in the sensor coordinate system, and $\{x''_{ij}, y''_{ij}, z''_{ij}\}$ is the corresponding transformed data set. Since there may be noises in the data acquisition process, a straight-through filter is used to perform basic outlier removal.

*B. Robust Target Region Segmentation*

The goal of the target region segmentation is to extract point cloud data lying on circular cross section. Currently, GMM is widely used in modeling segmentation models. In general, the standard GMM method assumes that the spectral measure of each independent ground object follows the Gaussian distribution, and uses the weighted average of the distribution of all ground objects to express the probability density function of the whole data set. Expectation Maximization (EM) algorithm is used to estimate the parameters of Gaussian distribution of each independent feature. In summary, according to the above analysis, for well applying GMM-based segmentation model, target region must satisfy the following two RULEs:

RULE IV. Target point clouds must be lying on circular cross section.

RULE V. The point clouds of each category on the axial profiles are independent, and belong to the same distribution.

*1) Analysis of Target Region Selection*

As shown in Fig. 9, the profile of a cross section for twist drill is composed of three parts, i.e. blade back, blade lip and blade groove, which are respectively marked with red, blue and green. The profile formed by blade back is located on a circle, which meets RULE IV. For our proposed measurement mode based on triangulation, it performs data collection by receiving light reflected from object. It subjects to three constraints: 1) measuring points must be located within the camera's depth of field range (DOF), 2) measuring points are within the field of camera vision (FOV), and 3) the angle between the reflected light of the measuring point and its surface normal vector must be less than the incident angle. So, during the scanning process, due to the shape characteristics of twist drill and the limitation of optical constraints, only blade back and part of blade lip can be scanned, which is shown in Fig. 10 (blue lines for blade back, red lines for blade lip and few part of blade groove). Therefore, the point cloud data of a drill for one scanning period is shown in the Fig. 11 (a). A set of two-dimensional data of an axial contour at the black line of the Fig. 11 (a) is shown in Fig. 11 (b). The *z* values of the point clouds presents a ladder distribution, where the upper part is the blade lip and the lower part is the blade back. The statistical histogram of the *z* values in a certain region is as shown in Fig. 11 (c). The depth values *z* of the point clouds jointly constitute a GMM, which meets RULE V.

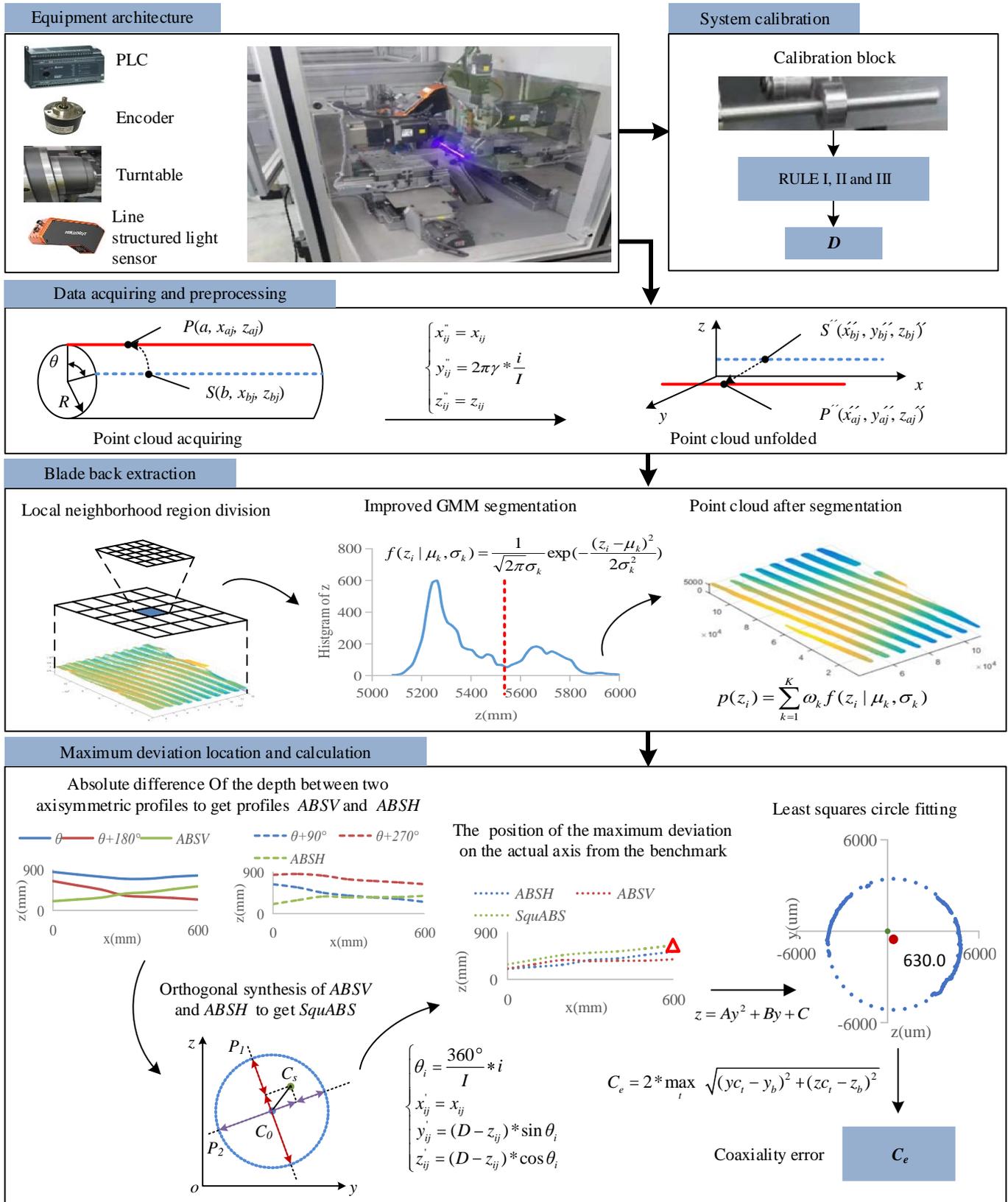

**Fig. 7.** Flow chart of our measurement framework.

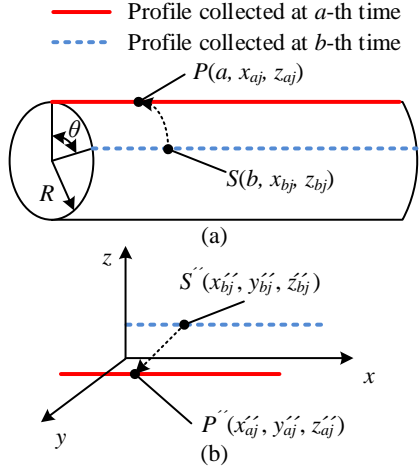

**Fig. 8.** Point cloud transformation. (a) Point cloud before transformation. (b) Point cloud after the transformation. *S* and *P* are the points respectively collected at the *a*-th and *b*-th time sampling. *S″* and *P″* are the points corresponding to *S* and *P* after the transformation respectively.

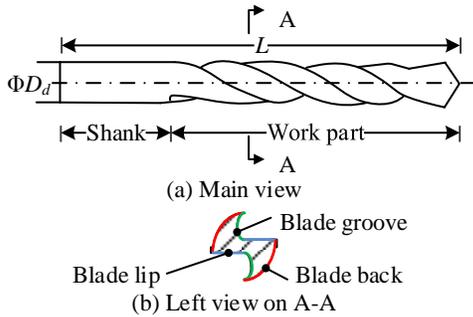

**Fig. 9.** Geometric figure of twist drill. Blade back is red, blade lip is blue, and blade groove is green.

*2) Homogeneous Region Model*

We use GMM to describe the depth values *z* of our point cloud data. In this model, the point clouds of blade back are defined as the foreground, and others are the background. So the density function $p(z_i)$ of a point $z_i$ would be defined as:

$$p(z_i) = \sum_{k=1}^{K} \omega_{ik} f(z_i \mid \mu_k, \sigma_k) \quad (6)$$

Where, $i=1, 2, ..., N$, $N$ is the number of point clouds, $k = 1, 2, ..., K$, $K$ is the number of the classes, $\omega_{ik}$ represents the weight of data $z_i$ belonging to the class $k$, with a constraint condition as follows:

$$0 \leq \omega_{ik} \leq 1 \text{ and } \sum_{k=1}^{K} \omega_{ik} = 1 \quad (7)$$

$f(z_i|\mu_k, \sigma_k)$ is the Gaussian distribution named a component of mixture model and specifically it denotes as:

$$f(z_i \mid \mu_k, \sigma_k) = \frac{1}{\sqrt{2\pi}\sigma_k} \exp(-\frac{(z_i - \mu_k)^2}{2\sigma_k^2}) \quad (8)$$

Where $\mu_k$ and $\sigma_k$ are the expectation and the standard deviation of the *k*-th Gaussian distribution.

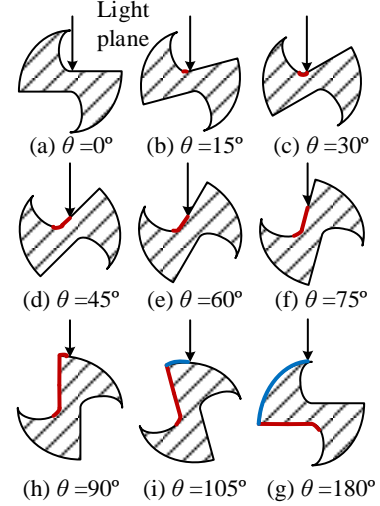

**Fig. 10.** Scanning process of drill cross section after certain rotation angle. Blue line is for blade back, red line is for blade lip and few parts of blade groove.

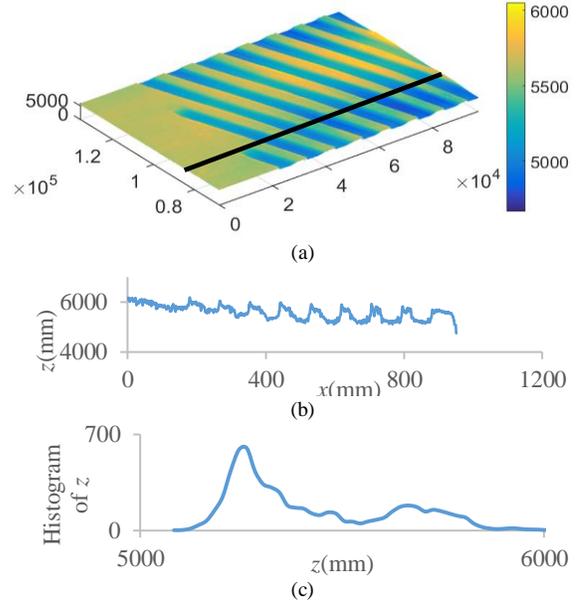

**Fig. 11.** (a) Point cloud of a drill. (b) The 2-D *x* and *z* profile data on black line in (a). (c) Histogram of *z* of a certain region on work part in (a).

Although the above method is simple and easy to implement, it has the following shortcomings: 1) The standard GMM method considers that points are independent from each other and does not consider the neighborhood effect of points, so the segmentation result is sensitive to noise. 2) This method represents the depth characteristics of a single type of region as a single-peak Gaussian distribution, but it is not ideal for fitting and merging the depth of point cloud data, especially high resolution data collected by line structured light sensor (homogeneous areas show significant multi-peak distribution due to line structured light measure difference).

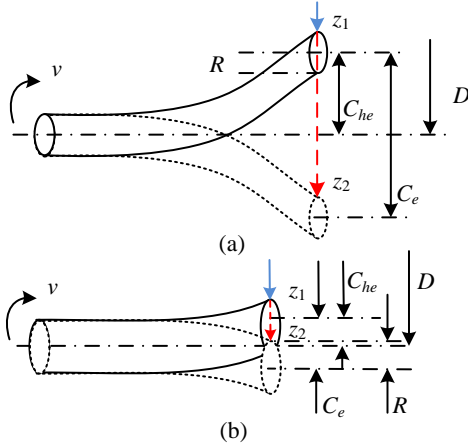

**Fig. 12.** Coaxiality approximate axisymmetric difference illustration. (a) Coaxiality is less than the diameter of the measured object. (b) Conditions of otherwise (a).

*3) Segmentation Decision Model*

In order to overcome the sensitivity of GMM to noise, we introduce local (neighborhood) spatial information, so that the category of each point is not only related to its own depth but also affected by the adjacent points. Based on the homogeneous region model, according to the principle that the category of any point in depth space is determined jointly by the probability that the point and its neighborhood points belong to this category, the segmentation decision model integrating spatial relations is established as following. A two-level division strategy is designed to perform local (neighborhood) spatial information extracting. Details are as following:

1) Point clouds are equally divided into $M*N*T$ blocks along the three axes, i.e. $O_CX_C$, $O_CY_C$ and $O_CZ_C$, and obtain the blocks set $B$, $b_i \in B$, $i=1, 2, ..., M*N*T$. $M$, $N$ and $T$ are the number of blocks along the $O_CX_C$, $O_CY_C$ and $O_CZ_C$ axes respectively. $T$ is set as 1.
2) Each block $b_i$ is evenly divided into $m*n*t$ patches along the three axes, i.e. $O_CX_C$, $O_CY_C$ and $O_CZ_C$, and obtain the patches set $F$, $f_j \in F$, $j=1, 2, ..., m*n*t$. $m$, $n$ and $t$ are the number of patches along the $O_CX_C$, $O_CY_C$ and $O_CZ_C$ axes respectively. $t$ is set as 1.
3) The statistical histogram for the depth value $z$ of every patch $h_j(z)$, is calculated, and the highest frequency value $zf_j$ can be calculated according to (9):

$$zf_j = \arg\max_{z_k}\{h(z)\}, k=1,2,...,\tau \quad (9)$$

$\tau$ is the number of point clouds within a patch $f_j$.

4) We initialize the parameters of the GMM as following:

$$\begin{cases} \sigma_F = \sigma_B = (\max\{zf_j\} - \min\{zf_j\})/4 \\ \mu_F = \sum_{j=1}^{m*n*t} zf_j - \sigma_F \\ \mu_B = \sum_{j=1}^{m*n*t} zf_j + \sigma_B \end{cases} \quad (10)$$

Where, $\mu_F$, $\sigma_F$, $\mu_B$ and $\sigma_B$ are the expectation and standard deviation of the GMM corresponding to the foreground and background respectively. $\omega_F$ and $\omega_B$ are set to be 0.5.

5) The maximum likelihood method (EM) is utilized to solve the GMM model, and achieve the classification of each patch $f_j$.

Furthermore, a statistical filter SOR is utilized for filtering the segmented point clouds in consideration of the inevitable noise, which is described in the experiments in detail.

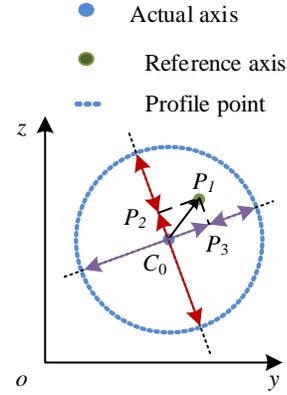

**Fig. 13.** Orthogonal decomposition and synthesis of coaxialitys.

*C. Rapid Axis Reconstruction and Coaxiality Calculation*

A rapid axis reconstruction method based on orthogonal synthesis is proposed. In the method, maximum deviation position of the axis is pre-positioned by using the extracted blade back data and the approximately reconstructed axis. The basis are the two characteristics of coaxiality, i.e. coaxiality is proportional to the opposite phase differences and neighboring phase differences are independent to each other. The so-called opposite phase difference is the absolute difference between any two axisymmetric profiles, which is proportional to the coaxiality. The so-called neighboring phase differences mean any two orthogonal axisymmetric profiles differences. So reversely, coaxiality can be orthogonal synthesized by any two orthogonal axisymmetric profiles differences.

*1) Analysis of Axis Deviation*

As for an part with good roundness and coaxiality, during the rotation around its axis, absolute depth difference between the axial contour under any angle $\theta$ and its axisymmetric contour at $\theta+180°$ is approximately 0. When the coaxiality increases, the contrast difference improves. Two cases of coaxiality in vivo and not in vivo are illustrated in Fig. 6 respectively. $R$ is the radius of twist drill, $C_e$ is the coaxiality and $C_{he}$ is half of $C_e$. In the Fig. 6(a), $z_1$ and $z_2$ are respectively the depth values of the same measuring points at any rotation angle of the drill $\theta$ and $\theta+180°$, respectively, which satisfies:

$$z_1 \approx D - R - 0.5*C_e \quad (11)$$

$$z_2 \approx D + (0.5*C_e - R) \quad (12)$$

According to (11) and (12), we can obtain:

$$C_e \approx z_2 - z_1 \quad (13)$$

Similarly, the situation in the Fig. 12(b) satisfies (14) and (15).

$$z_1 \approx D - (R + 0.5*C_e) \quad (14)$$

$$z_2 \approx D - (R - 0.5*C_e) \quad (15)$$

We can also obtain (13) according to (14) and (15).

The above deviation proves that axisymmetric difference can directly reflect the size of coaxiality. The coaxality can be orthogonal synthesis by two orthogonal axisymmetric profiles differences. The graphical illustration is as shown in Fig. 13. For any group of orthogonal directions $\vec{C_0P_2}$ and $\vec{C_0P_3}$, the axial deviation $\vec{C_0P_1}$ can be generated by orthogonal synthesis, which can be obtain from (16).

$$|\vec{C_0P_1}| = \sqrt{|\vec{C_0P_2}|^2 + |\vec{C_0P_2}|^2} \qquad (16)$$

Table I
MV-DP090-02B parameters

| Parameter | Value |
|---|---|
| Single contour points | 1920 |
| Measurement range (X-axis (width))-near side (mm) | 80 |
| Measurement range (X-axis (width))-far side (mm) | 153 |
| Measurement range (Z-axis (height))-near side (mm) | 106.5 |
| Measurement range (Z-axis (height))-far side (mm) | 200.0 |
| Resolution (X-axis) (mm) | 0.042~0.080 |
| Resolution (Z-axis) (mm) | 0.013~0.047 |
| Repetition precision (Z-axis)(um) | 3 |
| Scanning frequency (Hz) | 60~1000 |

Table II
Turntable parameters

| Parameter | Value |
|---|---|
| Angle resolution | 0.001° |
| Repeated positioning accuracy | 0.002° |
| Eccentric(um) | 5 |

*2) Location and Calculation of Maximum Deviation for Axis*

We use the orthogonal synthesis of any two groups differences of axisymmetric profiles to reconstruct the actual axis. The specific algorithm is as follows:

1) Two groups of orthogonal axisymmetric profiles are selected to contribute a point clouds set $C_i = \{x_{ik}, z_{ik}\}$, $k$=1, 2, ..., $K$, $K$ denotes the number of point clouds on every profile. $i$=1, 2, 3, 4, $i$ corresponds $\theta$, $\theta$+90°, $\theta$+180° and $\theta$+270° angle.
2) Quadratic spline functions of the selected profiles are constructed by utilizing the extracted point data.
3) The absolute differences of every two axisymmetric are calculated, the absolute difference of $\theta$ and $\theta$+180° are named as $ABSV$, the other two is denoted as $ABSH$, $ABSV=\{x_{hk}, z_{vk}\}$ and $ABSH=\{x_{hk}, z_{hk}\}$, which are calculated according to (17) and (18).

$$\begin{cases} x_{hk} = x_{0k} = x_{2k} \\ z_{hk} = |z_{0k} - z_{2k}| \end{cases} \qquad (17)$$

$$\begin{cases} x_{vk} = x_{1k} = x_{3k} \\ z_{vk} = |z_{1k} - z_{3k}| \end{cases} \qquad (18)$$

4) The approximate axis deviation profile is reconstructed by utilizing (19) and is denoted as $SquABS = \{xs_k, zs_k\}$.

$$\begin{cases} xs_k = x_{hk} = x_{vk} \\ zs_k = \sqrt{z_{hk}^2 + z_{vk}^2} \end{cases} \qquad (19)$$

5) The maximum deviation position of the actual axis is located by (20):

$$zs_{max} = \max\{zs_k\}, k = 1, 2, ..., K \qquad (20)$$

To reduce the location error, a threshold $\Delta zs$ is set, and the points whose $zs_k$ value satisfy (21) are collected to form a position set $XSM=\{xsm_t|t=1, 2, ..., T\}$, $T$ is the number of the cross sections positions.

$$zs_k \geq zs_{max} - \Delta zs \qquad (21)$$

6) Equation (1) and the set $XSM$ are used to obtain the point clouds set $CS$ of cross sections from the segmented blade back data. Where, $CS = \{(xsm_{tu}, ysm_{tu}, zsm_{tu})| t=1, 2, ..., T, u = 1, 2, ..., U\}$, $U$ is the number of the point cloud data on each cross section. Then we use the least-square circle method to fit the center of each cross section in set $CS$, and obtain a set $AS$ of axes. Where $AS = \{(yc_t, zc_t) | t = 1, 2, ..., T\}$. Similarly, we use the same method to fit the axes of the benchmark, and get the axes ($y_b$, $z_b$). Then, the coaxiality $C_e$ is following:

$$C_e = 2 * \max_t \sqrt{(yc_t - y_b)^2 + (zc_t - z_b)^2} \qquad (22)$$

IV. EXPERIMENTS AND DISCUSSION

*A. Setup*

The experimental point clouds were collected by a line structured light sensor named MV-DP090-02B. For the convenience of calculation, a region of interest is set up for the sensor, which makes that a single profile includes 1350 points. The specific parameters of the sensor are shown in Table I. Encoder is differential and the parameters of the turntable are detailed in Table II. In addition, the point cloud processing was performed on a computer with the windows 7, i7 CPU and memory 8G using a point cloud framework PCL1.8.1.

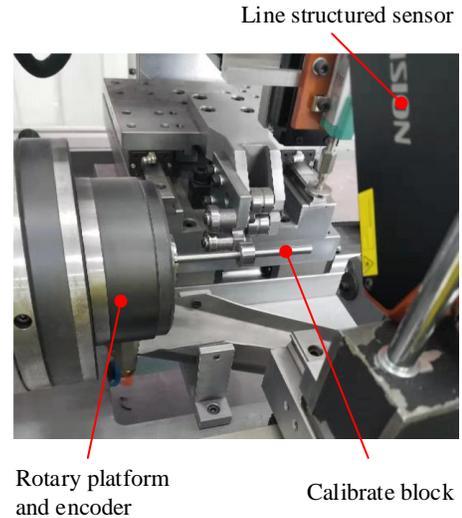

Fig. 14. The calibration of measuring equipment.

*B. System Calibration*

The calibration block as shown in Table III is machined with accuracy 2 um. Firstly, the device is fixed and adjusted to make the calibration block merge in the field of the line structured sensor vision, as shown in Fig. 14. Secondly, roll angle of the sensor is adjusted to make that the light plane of the sensor is parallel to the axis of the calibration block, and $\Delta z_{th}$ in (2) is set to be 1 um. Then the position of the sensor in the Y-axis direction is fine-tuned to let to be closest to the camera. By now the optimal installation position is reached, $z_{Bi}$ and $z_{Ci}$ is measured and parameter $D$ can be obtained. The point cloud of calibration block for one revolution is as shown in Fig. 15.

*C. Analysis of Blade Back Extraction*

In blade back extraction stage, the block size of the two-level division strategy have some influence on segmentation effect. A mass of experiments have been carried out, and show that the best performance were achieved when the size of the block was close to the sum width of single blade back and blade lip. The reason is that, the distribution difference between blade back and blade lip is the most obvious and the model is more convenient for convergence when the size is taken as that. The comparison of different block size for the two-level division strategy after blade back segmentation by our GMM-based method is shown in Fig. 16. The blade back region is red, and others are blue.

Table III
The calibration block parameter

| Parameter | Value |
| --- | --- |
| $l_a$(mm) | 120 |
| $l_b$(mm) | 70 |
| $l_c$(mm) | 10 |
| $l_d$(mm) | 40 |
| $\Phi d_a$(mm) | 8 |
| $\Phi d_b$(mm) | 14 |

In addition, to achieve more precise segmentation, other algorithms are attempted for comparative analysis, such as Region Growing (RG), Conditional Euclidean Distance (CE), Model Based (MB), and classical GMM [26] [27]. RG method assumed the blade back area as a plane, selected seeds randomly and took the normal vector threshold of neighborhood as the termination condition. CE method combined the intensity, normal vector and point cloud depth for extracting blade back region. MB method utilized the point cloud of drills to create model and recognized the blade back. The segmentation results are shown in Fig. 17.

As can be seen from Fig. 17, our method achieves the best segmentation result. Due to the rough and reflective surface of drills, MB method occurred large probability deviation in the estimation of the splines. Similarly, RG showed dreadful ability because of small differences between normal vectors owing to optics constraints. CE reaches the worst result.

*D. Analysis of Maximum Deviation Location for Axis*

For illustrating the effect, four drills with different bending degree are presented as shown in Fig. 18. Point cloud after blade back extraction is shown in Fig. 19. The point cloud data of four contours such as $\theta$, $\theta+90°$, $\theta+180°$ and $\theta+270°$ were selected for maximum deviation location after blade back segmentation. According to maximum axis deviation location method defined in section 3.3.2, the absolute difference value *ABSV* between $\theta$ and $\theta+180°$ and *ABSH* between $\theta+90°$ and $\theta+270°$ were calculated respectively, as shown in Fig. 20(a)(b)(d)(e)(g)(h)(j)(k). Then the actual axis of drills is approximately constructed by using orthogonal synthesis, as shown in Fig. 20(c)(f)(i)(l). Hollow △ is the position of the maximum axis deviation. As can be seen from Fig. 20(c) and (l), when two pairs of axisymmetric profiles differences are nearly the same, the axis deviation to the benchmark is larger than both of them. When two pairs of axisymmetric profiles differences differ largely, the axis deviation to the benchmark is closer to the larger one, as shown in Fig. 20(f) and (i). That is conforming to the rules of the orthogonal synthesis.

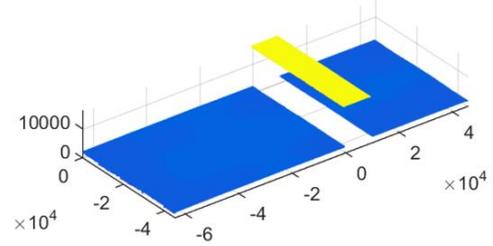

**Fig. 15.** The point cloud of calibration block on reaching RULE III.

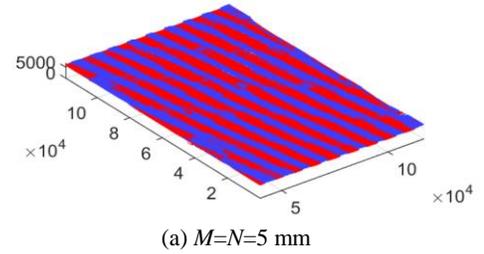
(a) *M=N*=5 mm

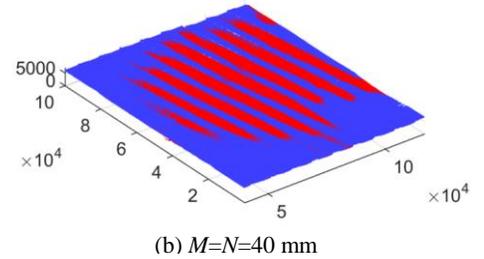
(b) *M=N*=40 mm

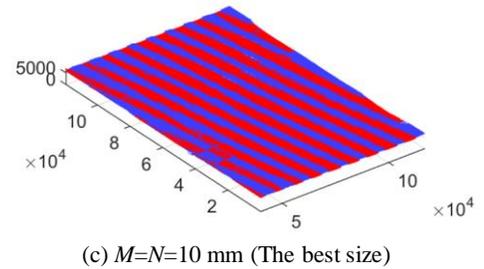
(c) *M=N*=10 mm (The best size)

**Fig. 16.** The comparison of different block size for the two-level division strategy after blade back segmentation by our improved GMM-based method.

According to the criterion of minimum containment area, coaxiality of drills is 2 times of radial distance from the center of cross sections on maximum axis deviation position to that of the benchmark. Here, the benchmark is on the shank. By utilizing the least-squares circle fitting method, the center of the benchmark and the cross section with the maximum deviation were calculated, as shown in Fig. 21. The points of the located cross section with maximum deviations are blue, and its center is red. The fitting center of the benchmark cross section is green.

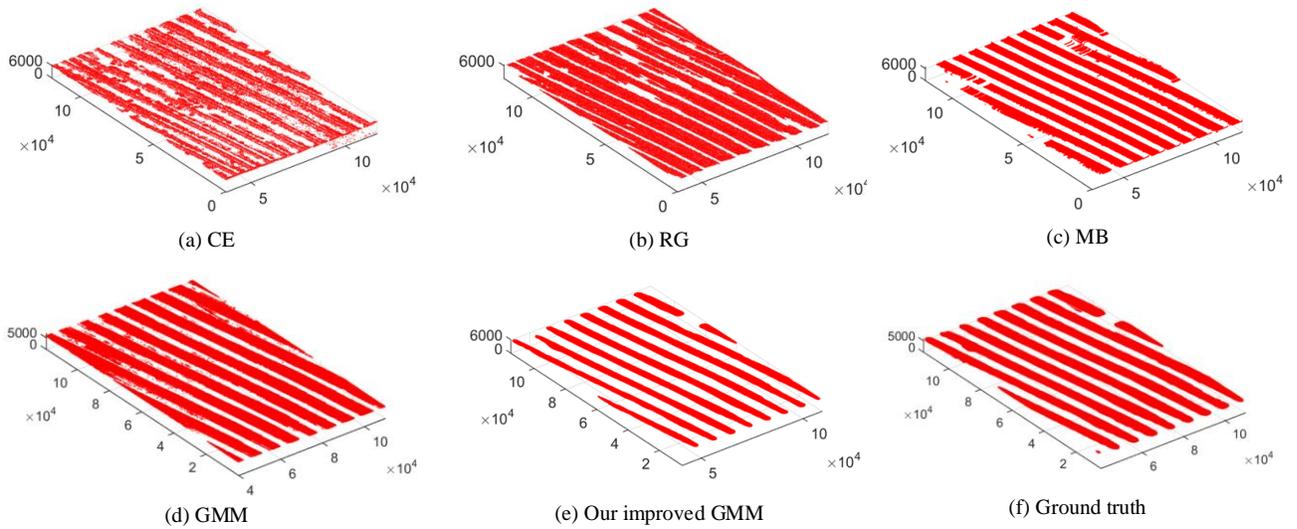

(a) CE  (b) RG  (c) MB  (d) GMM  (e) Our improved GMM  (f) Ground truth

**Fig. 17.** The comparison of different segmentation method.

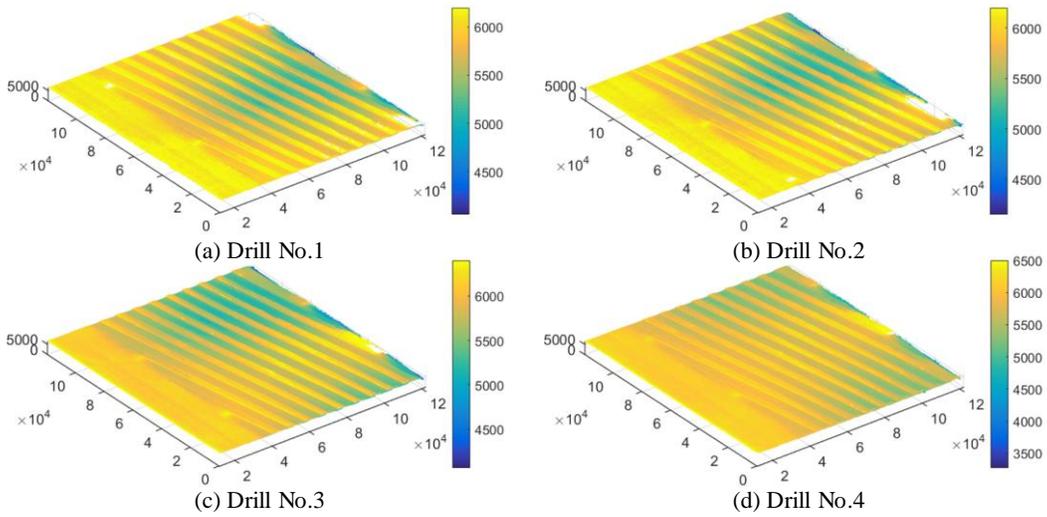

(a) Drill No.1  (b) Drill No.2  (c) Drill No.3  (d) Drill No.4

**Fig. 18.** Point cloud of Drill No.1~4 with different bending degree.

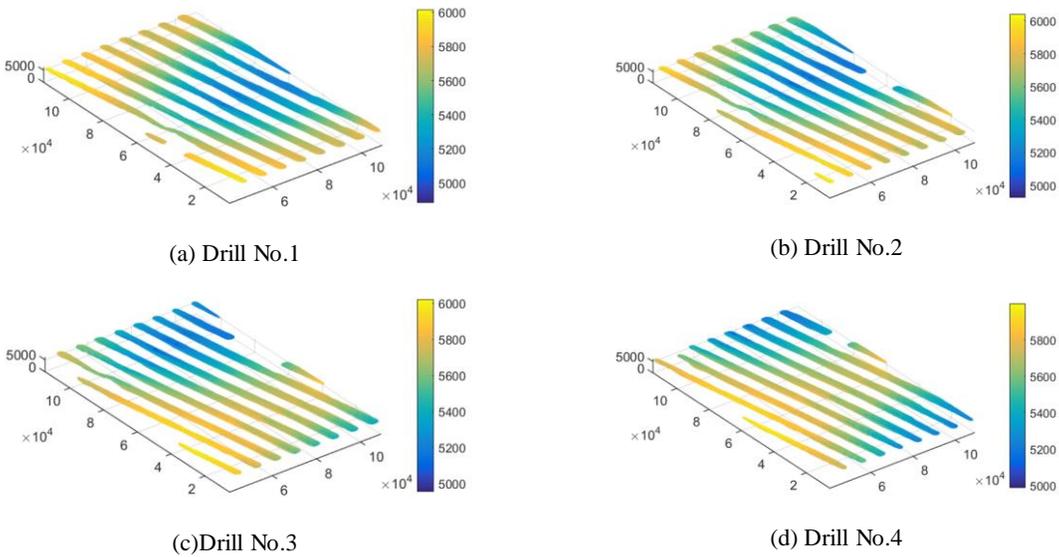

(a) Drill No.1  (b) Drill No.2  (c) Drill No.3  (d) Drill No.4

**Fig. 19.** Point cloud of blade back.

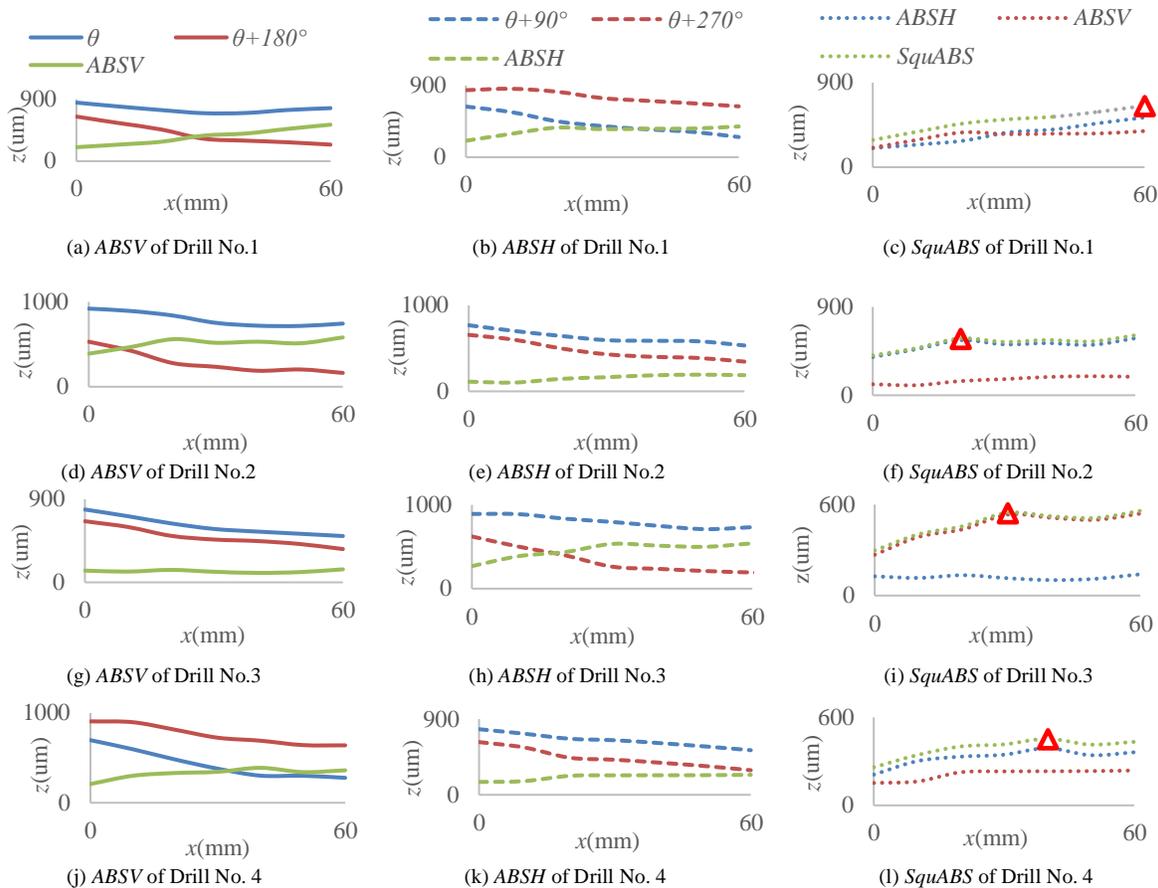

**Fig. 20.** The fitting curve of drills (No. 1-4) after rotating by $\theta°$, $\theta°+90°$, $\theta°+180°$, $\theta°+270°$ and the fitting curve of *ABSV*, *ABSH* and *SquABS*. (a) - (c) are the profiles of drill No. 1. (d) - (f) are the profiles of drill No. 2. (g) - (i) are the profiles of drill No. 3. (j) - (l) are the profiles of drill No. 4. Hollow triangle △ is the predicted maximum axis deviation position.

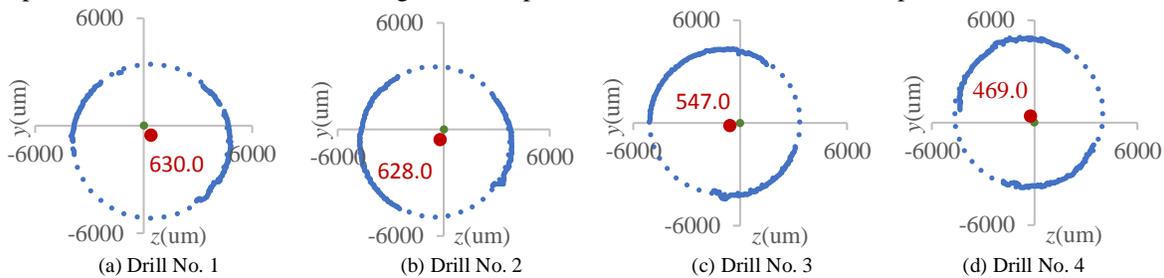

**Fig. 21.** Least-squares circle fitting for cross section. The points of cross section with maximum deviation are blue. The center of the fitting circle is red, and the center of the benchmark is green, the fitting circle of the cross section is blue.

Furthermore, an ocean of tests have been carried out on various specifications of twist drills. Thereinto, ten times measurement results of 4 drills with 100 mm in length and 10 mm in diameter are listed in Table IV corresponding Fig. 18. Manual method of V-type bracket plus dial meter and CMM method of ZEISS SPECTRUM were also adopted for comparing the accuracy and stability of measurements as shown in Fig. 22. It can be seen that our method performs a pleasurable result in accuracy and stability.

*E. Analysis of Measurement Uncertainty*

In the system, measurement accuracy is mainly affected by the precision and resolution of core components such as rotary table, differential encoder, and line structured light sensor. For the rotating-based measurement mode, the actual axis is obtained by measuring the radius $R$. Two kinds of displacements [36] may occur about the axis position including axial and radial displacement. Among them, the radial deviation will bring errors, while the axial deviation will not. Radically, the radial error is mainly introduced by turntable eccentricity and running accuracy. According to Table IIII, turntable eccentric is less than 5 um, so the uncertainty brought by which is 5 um recorded as $\Delta c$, and $\Delta c \leq 5$ um. The running angle error reacts worst when it occurs at the data acquiring moment, which generates the most deviation on the $O_wZ_w$ axis. The angel accuracy is 0.001° shown in Table II, so the uncertainty of running accuracy is $\Delta \eta = R*(1-\cos(0.001°))$, $\Delta \eta \ll \Delta c$, so it can be neglected. Otherwise, the precision of the line structure sensor is 3 um as shown in Table I, so its uncertainty is denoted as $\Delta z$, and $\Delta z \leq 3$ um.

Table IV
Measurement results for Drill No.1~4

| Drill | Method | 1th(mm) | 2th(mm) | 3th(mm) | 4th(mm) | 5th(mm) | 6th(mm) | 7th(mm) | 8th(mm) | 9th(mm) | 10th(mm) | Aver(mm) | Standard deviation |
|---|---|---|---|---|---|---|---|---|---|---|---|---|---|
| 1 | CMM | 0.621 | 0.633 | 0.635 | 0.626 | 0.625 | 0.631 | 0.623 | 0.636 | 0.629 | 0.627 | 0.629 | 0.005 |
|  | Manual | 0.790 | 0.450 | 0.675 | 0.536 | 0.680 | 0.710 | 0.540 | 0.750 | 0.690 | 0.240 | 0.605 | 0.167 |
|  | Ours | 0.626 | 0.639 | 0.629 | 0.627 | 0.635 | 0.631 | 0.629 | 0.635 | 0.623 | 0.627 | 0.630 | **0.005** |
| 2 | CMM | 0.620 | 0.623 | 0.624 | 0.622 | 0.627 | 0.625 | 0.628 | 0.629 | 0.621 | 0.624 | 0.624 | 0.003 |
|  | Manual | 0.670 | 0.710 | 0.620 | 0.640 | 0.610 | 0.420 | 0.640 | 0.530 | 0.540 | 0.650 | 0.603 | 0.084 |
|  | Ours | 0.619 | 0.625 | 0.630 | 0.621 | 0.635 | 0.626 | 0.622 | 0.631 | 0.633 | 0.639 | 0.628 | **0.007** |
| 3 | CMM | 0.546 | 0.543 | 0.544 | 0.543 | 0.547 | 0.545 | 0.543 | 0.544 | 0.546 | 0.544 | 0.545 | 0.001 |
|  | Manual | 0.770 | 0.550 | 0.560 | 0.690 | 0.550 | 0.540 | 0.430 | 0.570 | 0.530 | 0.520 | 0.571 | 0.094 |
|  | Ours | 0.543 | 0.554 | 0.546 | 0.549 | 0.550 | 0.551 | 0.544 | 0.553 | 0.539 | 0.540 | 0.547 | **0.005** |
| 4 | CMM | 0.465 | 0.462 | 0.467 | 0.469 | 0.468 | 0.470 | 0.466 | 0.472 | 0.469 | 0.468 | 0.468 | 0.003 |
|  | Manual | 0.650 | 0.470 | 0.360 | 0.570 | 0.660 | 0.490 | 0.450 | 0.570 | 0.470 | 0.460 | 0.515 | 0.095 |
|  | Ours | 0.468 | 0.467 | 0.463 | 0.473 | 0.469 | 0.466 | 0.465 | 0.467 | 0.472 | 0.475 | 0.469 | **0.004** |

In the calibration stage, the maximum system error occurs when the axis of rotating shaft is within light plane and generate maximum eccentricity, its uncertainty is denoted as $\Delta D=\Delta z+\Delta c$. In the measurement stage, the maximum system error occurs at the time that the center of drill cross section and the turntable axis are both within optical plane and meanwhile maximum eccentricity appears on the turntable. Then, the maximum deviation of the measured point is $\Delta z_p=\Delta z+\Delta c$. In short, the maximum deviation of measured radius is $\Delta R=\Delta D+\Delta z_p+\Delta \eta$. Referring to the tolerance range of radial jump variable of drills, the tolerance range of coaxiality is $C_s=0.03+0.01*(L/D_d)$. $L$ and $D_d$ are respectively the length and diameter of the measured drills. The proportion of the uncertainty of $R$ to the tolerance range of coaxiality is $\varepsilon$:

$$\varepsilon = \Delta R_{max} / 0.03+0.01*(L/D_d) \quad (23)$$

If the ratio is desired to be controlled within 20%, we can obtain:

$$R/L \leq 10 \quad (24)$$

This is fully compatible with the coaxiality measurement requirements for most specifications of twist drills. It confirms that, our devised mechanism and method are efficient and can be applied in most of the practical industrial scenes.

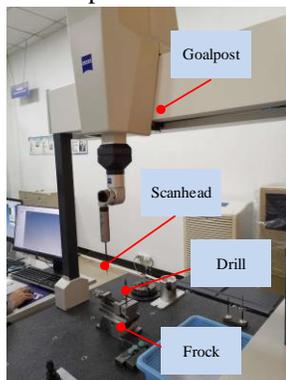

**Fig. 22.** CMM of ZEISS SPECTRUM.

## V. CONCLUSIONS

Since the twist drill is irregular rotating part with complex surface, it is challenging to achieve coaxiality measurement accurately and efficiently. In this paper, we propose an effective measurement mechanism of coaxiality for twist drill. Several core components are contained in the instrument, which can rapidly and accurately collect 3-D information of twist drill surface. Before formal measurement, a calibration process is conducted to get system parameter $D$ for compensating and eliminating system errors. In data processing stage, an improved GMM model is developed to segment collected point cloud data. To overcome the sensitivity of GMM to noise, a spatial neighborhood division strategy is designed, which can extract the point clouds data of the blade back accurately and robustly. To improve measurement efficiency, an axis reconstruction method based on orthogonal synthesis is designed. The novel method is accurate, and can rapidly locate the maximum deviation of the actual axis from the benchmark. In addition, the corresponding instrument is developed and experiments are carried out. The experimental results show that the measurement is effective with high precision (accuracy to 3 um) and high efficiency (measuring time < 3 s). It proves that the proposed mechanism can be applied in many other actual industrial scenes for online measurement, such as cylindricity, roundness or runout of a rotating body. Furthermore, the measurement uncertainty is analyzed, which verifies that the mechanism is suitable for most specifications of the twist drill.

In practice, several factors such as the resolution of the line structured light sensor, turntable precision or encoder subdivision degree would impact the measurement accuracy of coaxiality. It will be better to choose a higher-precision line structured sensor with finer laser line to improve the accuracy of the collected data. In addition, the measurement result may be influenced by frock abrasion or platform jitter of the hardware instrument. Several schemes could be considered to do, such as replacing with core components with higher resolution, adding the seismic platform, substituting frocks and the like, so as to achieve an ideal accuracy and reliability.